\documentclass[10pt,twocolumn,letterpaper]{article}
\usepackage[accsupp]{axessibility}  
\usepackage{cvpr}

\usepackage{graphicx}
\usepackage{amsmath}
\usepackage{ams symb}
\usepackage{booktabs}

\usepackage[pagebackref,breaklinks,colorlinks,bookmarks=false]{hyperref}

\usepackage[capitalize]{cleveref}
\crefname{section}{Sec.}{Secs.}
\Crefname{section}{Section}{Sections}
\Crefname{table}{Table}{Tables}
\crefname{table}{Tab.}{Tabs.}

\def\cvprPaperID{*****} 
\def\confName{CVPR}
\def\confYear{2022}

\begin{document}


 






\newpage

\title{Optimizing Nitrogen Management with Deep Reinforcement Learning  and \\Crop Simulations}  


\author{Jing Wu\thanks{These authors contributed equally.} \quad \quad Ran Tao\textsuperscript{$\ast$} \quad \quad Pan Zhao\textsuperscript{$\ast$}\quad \quad Nicolas F. Martin \quad \quad Naira Hovakimyan \\
University of Illinois at Urbana-Champaign\\
Urbana, IL 61801, USA \\
{\tt\small \{jingwu6, rant3, panzhao2, nfmartin,  nhovakim\}@illinois.edu}}

\maketitle

\begin{abstract}
Nitrogen (N) management is critical to sustain soil fertility and crop production while minimizing the negative environmental impact, but is challenging to optimize. This paper proposes an intelligent N management system using deep reinforcement learning (RL) and crop simulations with Decision Support System for Agrotechnology Transfer (DSSAT). We first formulate the N management problem as an RL problem. We then train management policies with deep Q-network and soft actor-critic algorithms, and the Gym-DSSAT interface that allows for daily interactions between the simulated crop environment and RL agents. According to the experiments on the maize crop in both Iowa and Florida in the US, our RL-trained policies outperform previous empirical methods by achieving higher or similar yield while using less fertilizers. 
\end{abstract}

\maketitle
\thispagestyle{empty}

\section{Introduction}
\label{sec: introduction}
The agricultural industry is facing significant challenges to meet the food demand to feed more than nine billion people by 2050 \cite{searchinger2019creating}. The challenges are further complicated by the current context of diminishing land and water resources, degraded soil and changing climate. The world urgently needs to move towards more sustainable and resilient cropping systems \cite{searchinger2019WRI-sustainable}. Among different factors influencing crop production and the environment, {\it nitrogen (N) management} is a key controllable one.   Nitrogen is the main nutrient affecting crop growth and yield formation, but excessive nitrogen has substantial
negative environmental effects \cite{sutton2011too-nitrogen}. 
Effective nitrogen management is therefore
crucial for  maximizing crop yields and
farmer income and minimizing negative environmental
impacts. 
Although best-practice knowledge for N management for common scenarios exists among farmers, it is unclear whether these practices are near-optimal, or whether some specific strategies transfer well to adverse seasonal conditions of extreme temperature or precipitation. 
N management is essentially a sequential decision making (SDM) problem as a few decisions on  nitrogen application time and quantities need to be made across the growth cycle of crops.
Modern reinforcement learning (RL) methods, represented by deep RL, have achieved remarkable or superhuman performance on a variety of tasks involving SDM such as gaming \cite{mnih2015human-deepRL,vinyals2019starcraft-rl}, data center cooling \cite{gamble_gao}, 
and robotic control \cite{kaufmann2018droneracing-rl,hwangbo2019agile-legged,song2021autonomous-deepRL}.  We expect that{ RL has a potential for optimizing agricultural management, improving the crop yield while minimizing the environmental impacts}. Training a deep RL policy often needs numerous interactions between the RL agent and the environment, which makes it unrealistic to leverage field trial-based approaches \cite{akkaya2019solving}.  Therefore, training the management policies in simulations  \cite{palmer2013influence,attonaty1997using,bergez2010design-crop-sim}, using crop models to simulate the crop and soil dynamics and interact with the RL agent, seems the only realistic solution. 
\begin{figure}[t]
  \begin{center}
    \includegraphics[width=0.49\textwidth]{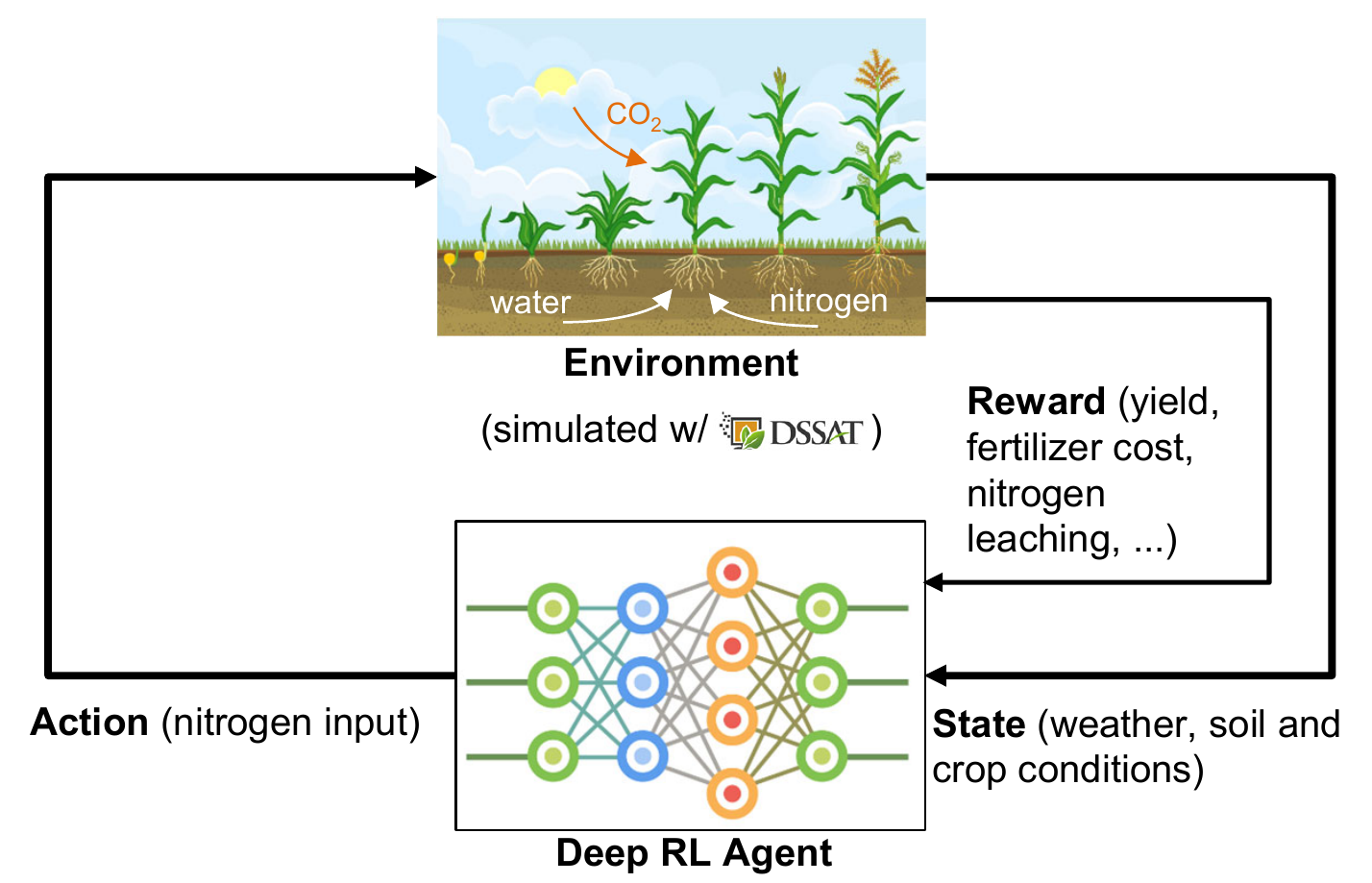}
  \end{center}
  \vspace{-3mm}
\caption{A framework for optimizing N management with deep RL and DSSAT-based crop simulations}\label{fig:framework}
\end{figure}

In this paper, we {\it propose and evaluate a framework for optimizing N management using deep RL and crop simulations}, depicted in Fig.~\ref{fig:framework}. In particular, we leverage Decision Support System for Agrotechnology Transfer (DSSAT), a widely used tool for crop modeling and simulation  \cite{jones2003dssat,hoogenboom2004dssat}, and the Gym-DSSAT interface \cite{gymdssat} that allows users to read the simulated crop and soil conditions and apply management practices on a daily basis. As a demonstration of the use of the presented framework, we train N management policies with two deep RL algorithms, namely deep Q-network (DQN) and soft actor-critic (SAC),  for the maize crop in Iowa and Florida, US. We further evaluate the performance of the trained policies in comparison with standard practices, and under different scenarios including partial observations and reduced action frequencies.

Compared to early work on RL-based crop management \cite{garcia1999rl-wheat,sun2017rl-irrigation}, our framework, which leverages deep RL, can handle much larger state and action spaces. Compared to recent work on deep RL-based agricultural management \cite{overweg2021cropgym,ashcraft2021ml-aided}, the crop model adopted in our framework, i.e., DSSAT, is much more widely used globally; additionally, our experimental study is significantly more comprehensive, which involves two different deep RL algorithms,  two geographic locations, and ablation study for partial observations and reduced action frequencies. 

\section{Related Work}
\subsection{Reinforcement Learning in Agricultural Management}
Reinforcement learning, as a sub-field of machine learning, aims to solve SDM problems by letting an agent directly interact with the environment and learn from trial and error \cite{Sutton2018RLintro}. As a pioneering work, \cite{garcia1999rl-wheat} proposed to use a simple RL method (namely, R-learning) and crop simulations to optimize management of wheat crops in France. \cite{sun2017rl-irrigation} studied the use of SARSA($\lambda$), an on-policy RL method,  and crop simulations to optimize the irrigation for the maize crop in Texas, US. 
However, the state and action spaces in \cite{garcia1999rl-wheat,sun2017rl-irrigation} were quite small due to the curse of dimensionality from which early RL methods suffered. For instance, the state space in \cite{sun2017rl-irrigation} has only one state, i.e., total soil water (TSW) level. In contrast, modern RL methods,  represented by deep RL, are able to handle extremely large state and action spaces due to the use of deep neural networks (DNNs) (to approximate the value functions or policies), and have achieved remarkable or superhuman performance on a variety of high-dimensional problems such as gaming \cite{mnih2015human-deepRL,vinyals2019starcraft-rl}, data center cooling \cite{gamble_gao}, 
and robotic control \cite{kaufmann2018droneracing-rl,hwangbo2019agile-legged,song2021autonomous-deepRL}. 
Deep RL based on the proximal policy optimization (PPO) algorithm  was used in \cite{overweg2021cropgym} to optimize the fertilizer management for the wheat crop. 
Additionally, \cite{ashcraft2021ml-aided} studied the use of PPO to optimize the irrigation management for russet potatoes. However, the study is quite coarse and the results are not promising. For instance, in terms of results, \cite{ashcraft2021ml-aided}   included only a simple learning curve showing the normalized reward, while the variables farmers mostly care such as  yield and management cost were not included. Additionally, the trained policy performed much worse than a simple policy which applies a constant amount of water.

\subsection{Crop Models}
Crop models can simulate crop growth in response to soil, water, nutrient, and weather dynamics. They are playing increasingly important roles in the development of sustainable agricultural management, because field and farm experiments require large amounts of resources and may still not provide sufficient information in space and time to identify appropriate and effective management practices \cite{jones2017brief-history-ag-models}. The development of crop models dated back to 1950s. In the past seven decades, many crop models of varying complexities have been developed by different groups, which include Agricultural Production Systems Simulator (APSIM), CERES (now contained in the DSSAT Suite of crop suite), CROPSYST, EPIC, WOFOST, and COUP. See the survey paper \cite{jones2017brief-history-ag-models} and a comparison of different crop models for yield response prediction \cite{salo2016comparing-crop}.
Among the existing crop models, the ones that are extensively used globally are APSIM and DSSAT, which are still constantly evolving and currently open-source to facilitate community-based development. 

Most of the existing crop models need the management practices to be pre-specified before the start of a simulation, while RL-based training of management policies requires  the management practices to be determined according to the soil, plan and weather conditions on a daily or weekly basis during the simulation. In light of this, the authors of \cite{overweg2021cropgym} developed the CropGym environment for training of  N management policies, which provides an interface to Open AI Gym \cite{brockman2016openai-gym}, a widely used toolkit for RL research, and enables an RL agent  to interact with the crop environment weekly.  However, CropGym is based on the LINTUL-3 model \cite{shibu2010lintul3} for the wheat crop, which has limited use. In a similar spirit, \cite{ashcraft2021ml-aided} presented another crop environment with the Open AI Gym interface for the russet potato  based on the SIMPLE crop model \cite{zhao2019simple-crop}, which, again, has limited use, potentially because the model is over-simplified. Recently, a Gym-DSSAT environment for the maize crop, which is based on the widely used DSSAT suite of crop models and provides a Gym interface, was developed \cite{gymdssat} and enables an RL agent to interact with the environment on a daily basis. However, there have been no results on the use of Gym-DSSAT for training crop management policies up to now. 
\section{Methods}
\label{section: methods}
We now present technical details for the N management framework based on deep RL and crop simulations depicted in \cref{fig:framework}. 
\subsection{MDP Problem Formulation}
The N management problem can be formulated as a finite Markov decision process (MDP) problem. In this formulation, a decision-making agent continuously interacts with the environment. At each day $t$, the agent selects an action (i.e., management practice), $a_t$, from the action space $\mathcal{A}$, based on the current state $s_t$, which is an array of elements from the state space $\mathcal{S}$. The selected action is applied to the environment and a new state ($s_{t+1}$) is generated based on this action; meanwhile, a reward signal $r_t = r(s_t,a_t)$ is produced to evaluate the immediate consequence of the selected action. This interaction repeats until the termination of the interaction, e.g., when the crop is harvested. The goal of the agent is to select optimal actions to maximize the future discounted return. The future discounted return at time $t$ is defined as $R_t = \sum_{\tau=t}^T\gamma^{\tau-t}r_\tau$, where $T$ is the time step at termination.

For N management, the action space $\mathcal A$  contains all possible amounts of nitrogen applied at a day. {All the states that compose the state space  are listed in Table~\ref{table:state-description}.} The reward function $r(s_t,a_t)$ at day $t$ is set as:
\begin{equation}\label{eq:reward}
    r(s_t,a_t) \!=\! \left\{\hspace{-2mm}
    \begin{array}{ll}
         w_1Y\!-\! w_2 a_t \!-\! w_3 N_{l,t} \!-\! w_4 P_t & \hspace{-2mm}\textup{if }  \textup{harvest at $t$,}    \\
        \!- w_2 a_t \!-\! w_3 N_{l,t} \!-\! w_4 P_t  & \hspace{-2mm} \textup{otherwise,} 
    \end{array}
    \right.
\end{equation}
where $a_t$ is the action (i.e. amount of nitrogen applied at day $t$), $N_{l,t}$ is the nitrate leaching at day $t$, $Y$ is the crop yield at the harvest date represented by the top weight at maturity, and $P_t$ is the additional penalty on large {\it total amount} of nitrogen applied. In particular, $P_t =  \sum_{k=1}^t a_k - {threshold}$ if $a_t \neq 0$ and $P_t = 0$ if $a_t  = 0$, where $threshold$ represents the allowable total amount of nitrogen inputs.  It may be worth mentioning that nitrate leaching occurs when nitrate is washed out of the root zone by heavy rainfall.  Leaching is undesirable because it leads to the waste of the fertilizers, and more importantly, causes environmental problems such as eutrophication of watercourses and soil degradation. Thus, we include a penalty on nitrate leaching in the reward function.
Finally, the positive constants  $w_1\sim w_4$ are selected to balance the different aspects mentioned above.  

\subsection{Training Management Policies using Deep RL}

For solving the formulated MDP problem, we leverage the recently proposed deep RL algorithms, which have achieved remarkable  performance on a variety of tasks  \cite{mnih2015human-deepRL,vinyals2019starcraft-rl, gamble_gao,jin2018real,kaufmann2018droneracing-rl,hwangbo2019agile-legged,song2021autonomous-deepRL}. We choose deep Q-network (DQN) \cite{mnih2015human-deepRL} and soft actor-critic (SAC) \cite{haarnoja2018soft} for the experimental study, but other deep RL algorithms capable of handling continuous state spaces can also be applied. 

\subsubsection{Policy Training with DQN}

DQN is a model-free value function based deep RL algorithm, which uses a deep neural network (DNN) to approximate the action-value function in Q-network \cite{mnih2015human-deepRL}. The essential idea of DQN is to learn an optimal action-value function $Q^\star(s,a) = \max_{\pi}\mathop{\mathbb{E}}[R_t|s_t=s,a_t=a,\pi]$, where $\pi$ is a policy mapping a state $s_t$ to an action $a_t$ at a given time $t$. With the $Q^\star$ function, given an action $s_t$, an optimal action $a_t^\star$ can be determined, e.g., by following a greedy policy defined by  $a_t^\star = \max_{a\in \mathcal{A}}Q^\star(s_t,a)$. From the interaction between the agent and environment, tuples of $(s,a,r,s')$ are generated and stored in a replay buffer, where $s,a,r$ and $s'$  denote current state, current action, immediate reward obtained by applying the action $a$ at the state $s$, and next state, respectively. Due to the nature of continuity of the action space, we discretized the action space. At iteration $i$, the Q network can be trained by minimizing the loss function: 
\begin{gather}
\hspace{-2mm}   L_i(\theta_i) \! \triangleq
  \!\! \mathop{\!\mathbb{E}}_{(s,a,r,s')}\!\left[r\!+\!\gamma \max_{a'\in \mathcal{A}}Q(s'\!,a';\theta_i^{-}) \!-\! Q(s,a;\theta_i)\!\right]\!,
\end{gather}
where the tuples $(s,a,r,s')$ are sampled from the replay buffer, $\theta_i$ are the parameters of the Q-network at iteration $i$, and $\theta_i^{-}$ are the network parameters used to compute the target at iteration $i$. The optimization problem can be solved using stochastic gradient descent algorithms \cite{mnih2015human-deepRL}. 

\subsubsection{Policy Training with SAC}
SAC is a policy-gradient deep RL algorithm  that represents the state of the art among model-free RL algorithms in terms of
sample efficiency and stability with respect to the hyperparameters \cite{haarnoja2018soft}. Besides the expected sum of rewards, SAC introduces the expected entropy to favor stochastic policies, which leads to a cost function  $L$ defined by \begin{gather}
    \label{eq:sac}
    L \triangleq
-\sum_{t=0}^{T}\mathbb{E}_{(s_{t},a_{t})\sim}p_{\pi}\left [ r(s_{t},a_{t})+\alpha \mathcal{H}(\pi(\cdot|s_{t})) \right ], 
\end{gather}
where $p$ denotes the state-action marginals of the trajectory distribution, $\mathcal{H}$ determines the entropy for the evaluation of randomness given the state $s_{t}$, and $r(s_{t},a_{t})$ is the immediate reward at time $t$. The temperature parameter $\alpha$ decides the trade-off between the entropy term and rewards.



\subsection{Simulating the Crop Response using Gym-DSSAT}

DSSAT has been used for various crop simulations worldwide in the last 30 years\cite{jones2003dssat}. However, limited interactions can be reached during the running period of simulation, leading to a possible delay of adjustment for management decisions. Recently, Gym-DSSAT\cite{gymdssat} has been developed to bridge the communication gap between the simulation environment and daily management decisions. This communication pipeline enables RL researchers to manipulate DSSAT like Open AI Gym in machine learning and robotics\cite{mnih2015human-deepRL,vinyals2019starcraft-rl}. In Gym-DSSAT, the environment is defined at a field scale with a time step corresponding to one day. An episode typically covers about 160 days from planting to harvest, and its state is automatically set as "done" at crop maturity. Weather is randomly generated via WGEN's\cite{richardson1985weather} built-in stochastic weather generator and can be fixed depending on simulation purposes. 

With Gym-DSSAT, millions of daily interactions between an RL agent and the simulated crop environment can be achieved in a few minutes, and used for training the management policies. 

\section{Experiments and Results}
\label{sec: experiments and results}
We conducted experiments on training N management policies for the maize crop in both Florida and Iowa. {These two locations are selected since they have different weather and soil conditions, which can be leveraged to test the general applicability of the proposed framework. Also, DSSAT includes templates for simulating the maize crop in these two locations, which facilitates the implementation of our proposed framework.} We evaluated the performance of trained policies in comparison with the standard practice proposed in \cite{mandrini2021understanding-nitrogen}. 
\subsection{Datasets for Florida and Iowa}
\label{sec:datasets}
Two experiments were studied. The first one is for the maize crop in Ames, Iowa, in 1999. The simulation starts on April 25th, the planting happens on May 27th, and  the crop is harvested no later than Oct 24th. The soil has a depth of 151 cm, and the plant density is 7.6 plant/m$^2$. The second experiment is for the maize crop in Gainesville, Florida, in 1982. In the Florida setup, the simulation starts on Jan 30th, while the crop is planted on Feb 26th and harvested when reaching maturity. The soil in this case has a depth of 180 cm, and the plant density is 7.2 plant/m$^2$. For both simulations, the irrigation is set to 0. This is consistent with the current practice in Iowa, where the maize crop is not irrigated. On the other hand, irrigation is crucial to improve the maize yield in Florida in reality. Setting the irrigation to 0 for the Florida case can be considered as an emulation of the extreme case of severe drought and limited water supply, which allows us to compare the RL-based management strategy with the standard practice under this extreme case.  

\subsection{Implementation Details}
For all the experiments, weight parameters $w_1, w_2$, and $w_3$ in the reward function \eqref{eq:reward} were set to be 0.1, and $w_4$ is set to be 1. For both DQN and SAC, we implemented the training using Pytorch, and used the Adam\cite{kingma2014adam} optimizer with an initial learning rate of 0.00005 
and a batch size of 64
to train the neural network. 
We trained the policies for 1200 episodes with the exploration rate $\epsilon$ decreasing from 1 to 0, following a decay factor of 0.994 for the Florida case and of 0.992 for the Iowa case. 

For DQN,  the discrete action space was defined to be $\mathcal A =\{40k \frac{\textrm{kg}}{\textrm{ha}}|k=0,1,2,3,4\}$. The discount factor was set to be 0.99.

For SAC, the agent action $a_{sac}$ varies from 0 to 200 and is discretized into the same action space as the one used for DQN through the mapping:
$	\mathop{\arg\min}_{a\in \mathcal A} \ \ \| \mathrm{} a_{sac}-a\|,
$ for both training and testing. The discretization is for being consistent with farmers' fertilization patterns, i.e., fertilize only a few times in the whole growth cycle. 
The discount factor and smoothing constant  for updating the target network were set to be 0.98 and 0.001, respectively. 

For comparison with the trained policies, we also implemented the standard management practice in \cite{mandrini2021understanding-nitrogen}, which suggests to add nitrogen at vegetative growth stage (vstage) 5, the stage when crop reaches five expanded leaves. 

\subsection{Results for the Iowa Maize}\label{sec:sub-results-Iowa}
\begin{figure}[t]
  \centering
  \includegraphics[width=0.93\linewidth]{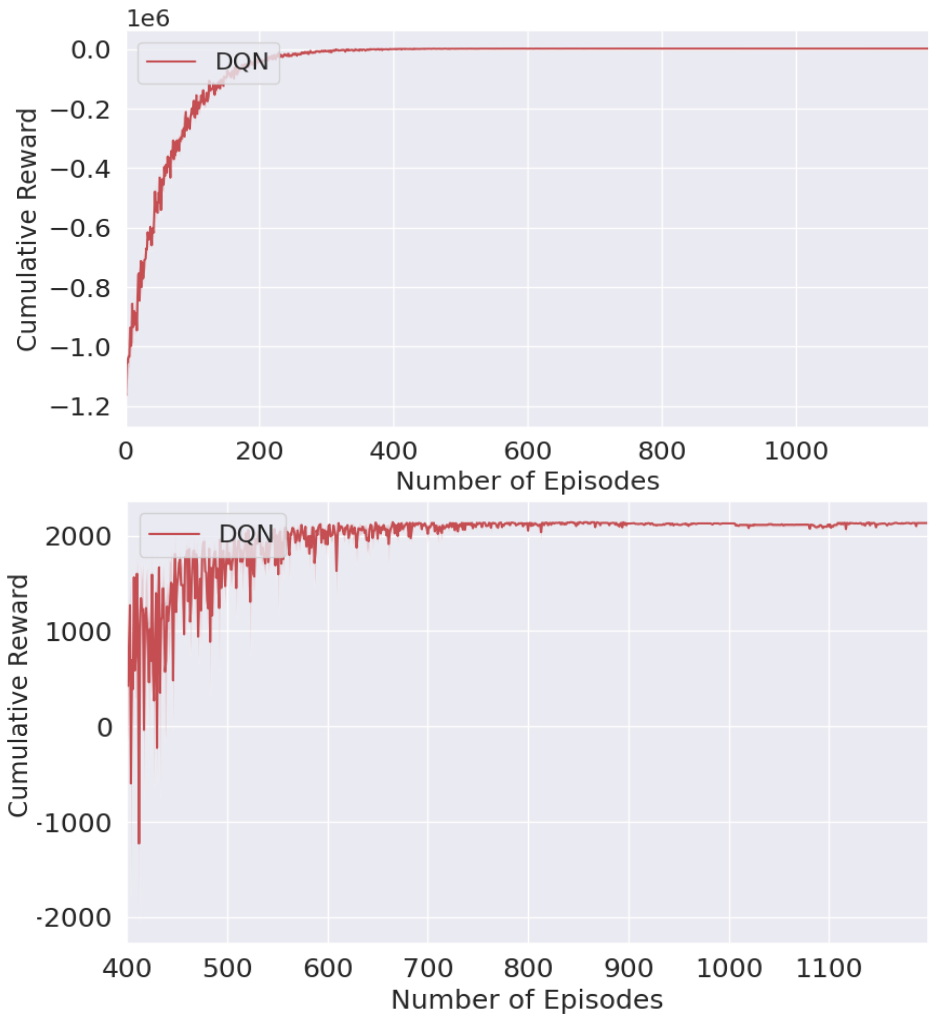}
   \caption{Cumulative reward versus episodes with DQN for Iowa. Results are averaged over five trials, with the light-red shaded area denotes the variance. Top: full view. Bottom: zoomed-in view for 400--1200 episodes}
   \label{fig: rewards of Iowa}
\end{figure}

The training curve using DQN, averaged over five trials, is shown in \cref{fig: rewards of Iowa}. During the first 200 episodes of exploration, due to the large exploration rate, the DQN agent over-fertilizes, causing significant penalties. After 800 episodes of training, the learning converges, constantly giving a cumulative reward of over 2000. Concretely, Table~\ref{table:Iowa} compares the performance of the DQN-trained policy and three baseline strategies, corresponding to 160, 240, 280 kg/ha of nitrogen applied at stage v5, as suggested in \cite{mandrini2021understanding-nitrogen}. The trained DQN agent decides to apply a total of 240 kg/ha nitrogen input during the growing season, and achieves 21711.8 kg/ha top weight of maize at maturity and a cumulative reward of 2126.3. Among three baseline policies, the one with 280 kg/ha achieves the largest cumulative reward of 2142.9 and largest top weight of 21709.5 kg/ha. Compared with the best baseline, DQN achieves slight improvement on top weight at maturity while using 14\% less nitrogen input, being more cost-efficient. Compared to the baseline using same amount of nitrogen input, DQN achieves a 1\% increment on the top weight at harvest. In general, the trained DQN agent achieves better results than the baseline methods.

The performance of SAC is shown in Table~\ref{table:Iowa_S}. Although using less nitrogen that causes a smaller top weight at maturity, the SAC policy still achieves a  cumulative reward similar to that achieved by the DQN policy. Thus, both RL algorithms succeeded in finding a better management policy than the baselines. However, the learning process converged much faster with SAC. Specifically, the cumulative reward with SAC reached 2100 within 700 episodes, while additional 300 episodes were needed to achieve similar results with DQN. 

\begin{table*}[!ht]
    \centering
    \footnotesize
    \caption{Performance  comparison between DQN and baseline policies for Iowa. Baseline (X) indicates that X kg/ha of nitrogen is applied at stage v5.}
    \begin{tabular}{ l|l l l l l }
    \hline 
        Methods & Nitrogen input (kg/ha) &  Nitrate leaching (kg/ha) & Nitrogen uptake (kg/ha) & Top weight at maturity (kg/ha)  & Cumulative reward \\ \hline \hline
        Baseline (160) & 160 & 0.11 & 219.1 & 21133.3 & 2097.3 \\ \hline
        Baseline (240) & 240 & 0.11 & 264.0 & 21502.9 & 2126.3 \\ \hline
        Baseline (280) & 280 & 0.11 & 272.3 & 21709.5 & 2142.9 \\ \hline
        DQN & 240 & 0.12 & 290.4 & \textbf{21711.8} & \textbf{2147.1} \\ \hline
    \end{tabular}
    \label{table:Iowa}
\end{table*}

\begin{table*}[!ht]
    \centering
    \footnotesize
    \caption{Performance comparison between SAC and DQN policies for Iowa.}
    \begin{tabular}{l|l|l|l|l}
    \hline
        Methods & Nitrogen input (kg/ha) & Top weight at maturity (kg/ha)  & Cumulative reward  &  Episodes of convergence \\ \hline \hline
        DQN & 240 & 21711.8 & 2147.1 & 1000 \\ \hline
        SAC & 200 & 21503.3 & 2144.2 & 700 \\ \hline
    \end{tabular}
    \label{table:Iowa_S}
\end{table*}

\subsection{Results for Florida Case}
\label{sec:sub-results-Florida}
\begin{figure}[ht]
  \centering
  \includegraphics[width=0.9\linewidth]{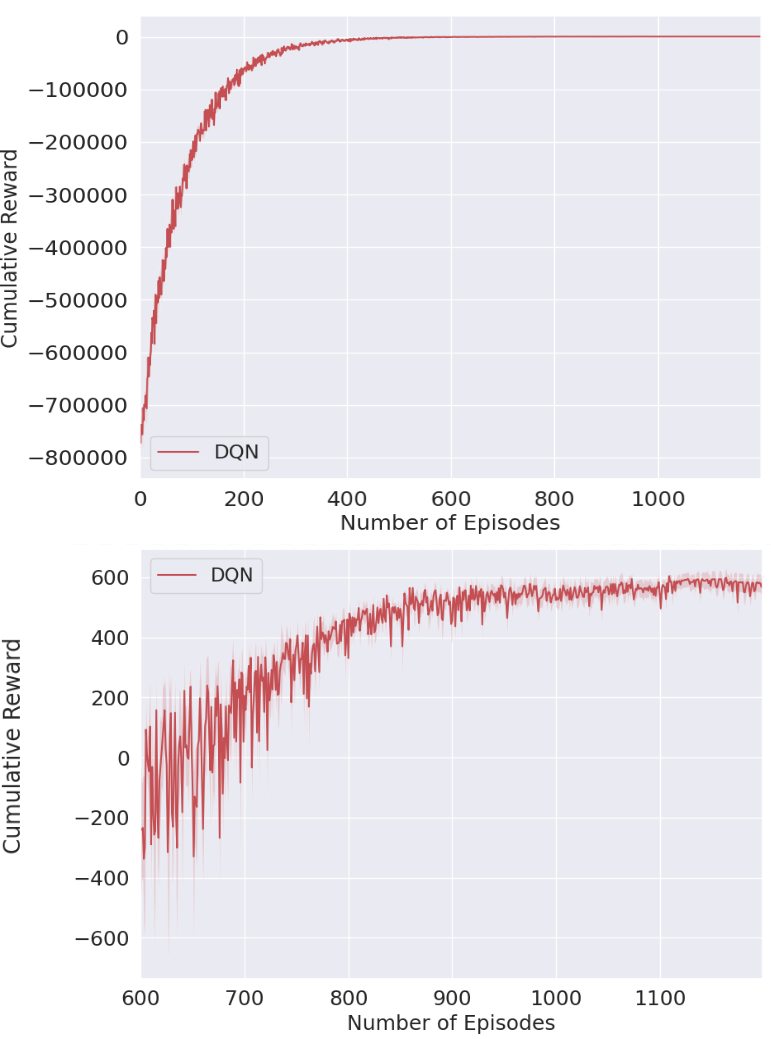}
   \caption{Cumulative reward versus episodes with DQN for Florida.  Results are averaged over five trials, with the light-red shaded area denotes the variance. Top: full view. Bottom: zoomed-in view for 400--1200 episodes}
   \label{fig: rewards of florida}
\end{figure}

 As we mentioned in~\ref{sec:datasets}, the Florida case is not realistic due to the 0 irrigation setup, and can be considered as an extreme weather case under severe drought with water shortage. Accordingly, the yields obtained under this setup are much smaller compared to those under the Iowa case. 
 The training curve under DQN averaged over five trials is shown in \cref{fig: rewards of florida}. The performance comparison between our trained DQN policies and baselines is summarized in Table~\ref{table:Florida}. As one can see, the DQN policy shows a stable improvement in terms of the top weight and rewards, which is consistent with the results for Iowa. The performance comparison between SAC and DQN is shown in Table~\ref{table:Florida_S}. Similar to the Iowa case, the SAC policy achieved a similar cumulative reward as the DQN policy but the training with SAC converged much faster.


\begin{table*}[!ht]
    \centering
    \footnotesize
    \caption{Performance comparison  between  DQN and baseline policies for Florida. Baseline (X) indicates that X kg/ha of nitrogen is applied at stage v5.}
    \begin{tabular}{ l|l l l l l }
    
    \hline 
        Methods & Nitrogen input (kg/ha) & Nitrate leaching (kg/ha) & Nitrogen uptake (kg/ha) & Top weight at maturity (kg/ha)  & Cumulative reward \\ \hline \hline
        Baseline (40) & 40 & 46 & 55 & 4393.3 & 430.7 \\ \hline
        Baseline (80) & 80 & 65 & 66 & 4673.1 & 452.8 \\ \hline
        Baseline (160) & 160 & 97 & 86 & 5190.4 & 493.3 \\ \hline
        DQN & 80 & 33 & 105 & \textbf{6310.8} & \textbf{619.7} \\ \hline
    \end{tabular}
    \label{table:Florida}
\end{table*}

\begin{table*}[!ht]
    \centering
    \footnotesize 
    \caption{Performance comparison between SAC and DQN policies for Florida.}
    \begin{tabular}{ l|l l l l l }
    \hline 
        Methods & Nitrogen input (kg/ha) & Top weight at maturity (kg/ha)  & Cumulative reward &  Episodes of convergence \\ \hline \hline
        DQN & 80 & 6310.8 & 619.7 & 900 \\ \hline
        SAC & 80 & 6308.0 & 610.2 & 700 \\ \hline
    \end{tabular}
    \label{table:Florida_S}
\end{table*}

\subsection{Ablation Study}
In practice, not all the states used in the training and testing of the management policies in the previous sections are accessible.  Additionally, from an economic perspective, farmers prefer to make decisions less frequently, e.g., weekly and biweekly, instead of daily. Therefore, in this section, we study the effect of full/partial observation and action frequencies on the performance of the proposed framework.


\subsubsection{Full vs. Partial Observation}\label{sec:sub-partial-obs}
\begin{figure}[ht]
  \centering
  \includegraphics[width=0.9\linewidth]{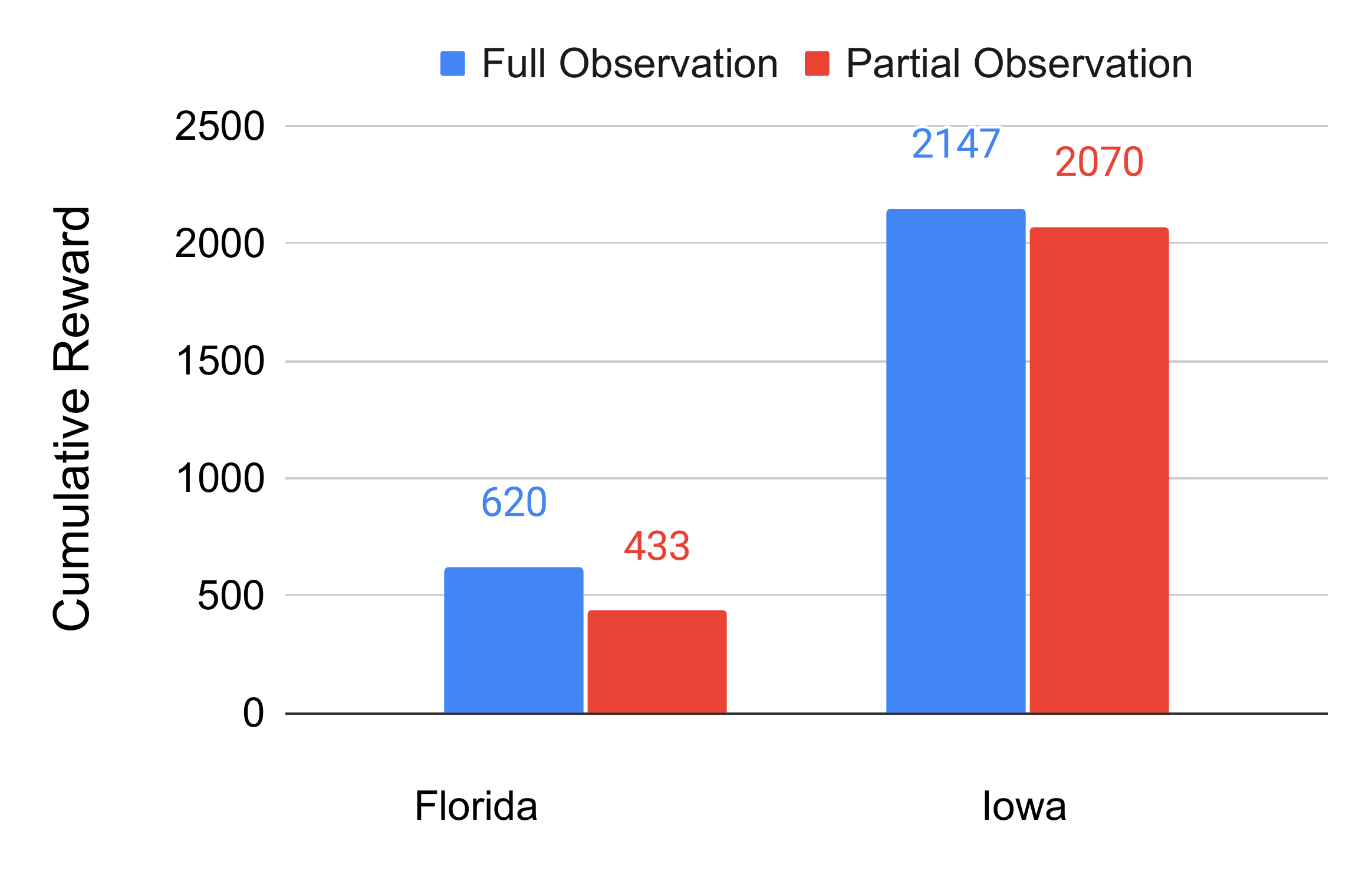}\\
    \includegraphics[width=0.9\linewidth]{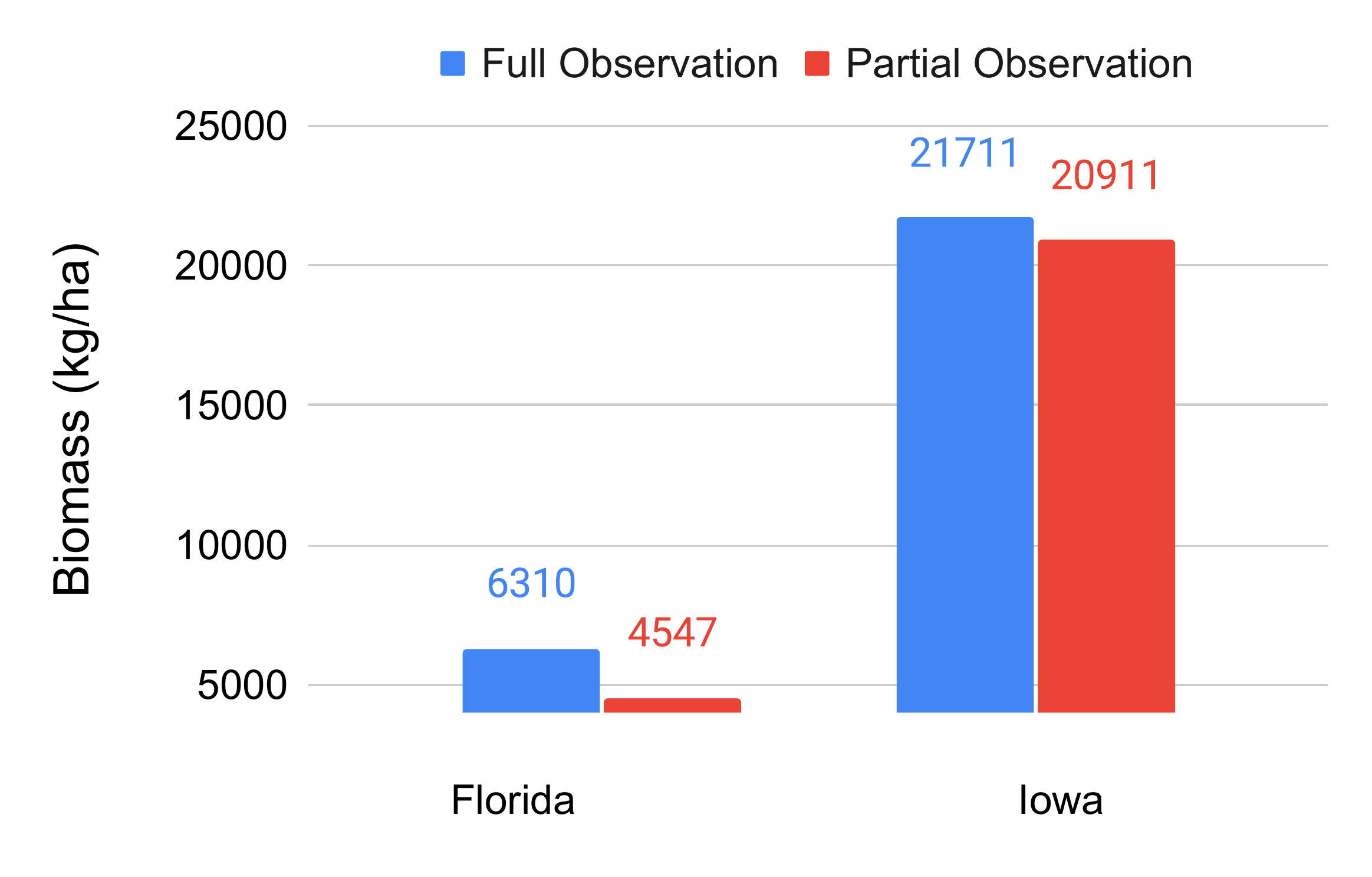}
   \caption{{Comparison of cumulative rewards (top) and final above the ground population biomass (bottom)} obtained under partial observation and full observation. The results are averaged over three trials.}
   \label{fig: patial observation}
\end{figure}

To understand the contribution of observed states in the fertilizer optimization process, we carry out an ablation study on  (i) full observation case, in which all the states listed in Table~\ref{table:state-description} are used for policy training and testing and (ii) partial observation case, in which only 10 states (indicated in Table~\ref{table:state-description}) are used. The study on partial observation is motivated by the fact that not all the states output by DSSAT can be accessed by farmers without professional agricultural tools for detection and inspection.  Experiments of full observation have been conducted in \cref{sec: experiments and results}. The results under partial observation, which are based on DQN, are shown in 
\cref{fig: patial observation}. For both Florida and Iowa, the policy training and testing under partial observation were conducted three times, and \cref{fig: patial observation}  shows the results averaged over the three trials.

For both Florida and Iowa setups, the policies under partial observation always underperformed those under full observation. In particular, we observed 30.15$\%$ and 3.58$\%$ decreases in reward and  27.94$\%$ and 3.68 $\%$ drops in the final yield for Florida and Iowa, respectively. {The decrease for Florida is relatively large compared to that for Iowa. This could be attributed to the extreme weather condition associated with Florida which requires the RL agent to have more comprehensive information to make a good decision.}


\subsubsection {Action Frequency}
We ablate the action frequency in the life cycle of maize to further understand the applied actions during the simulation process. We continue to conduct all the experiments with DQN in a discrete space to ensure the consistency with farmers' fertilization patterns. Concretely, we experiment with the trained DQN policy using two action frequencies: (i) RL agents are allowed to fertilize every day and (ii) RL agents are only permitted to fertilize every ten days. 
\begin{figure}[h]
  \centering
  \includegraphics[width=0.93\linewidth]{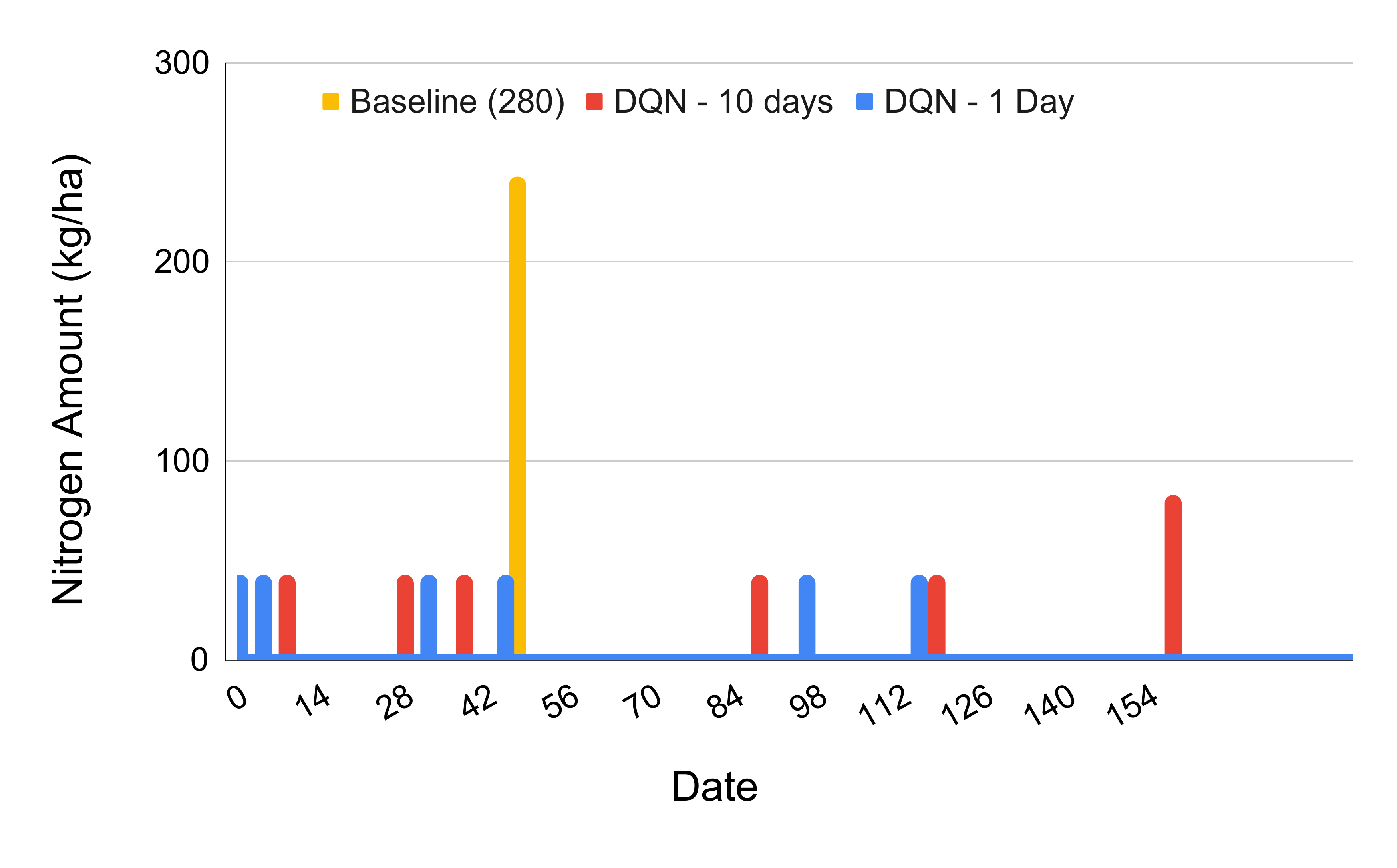}
   \caption{Applied actions under different action frequencies for Iowa.}
   \label{fig: action Iowa}
\end{figure}
\begin{figure}[h]
  \centering
  \includegraphics[width=0.93\linewidth]{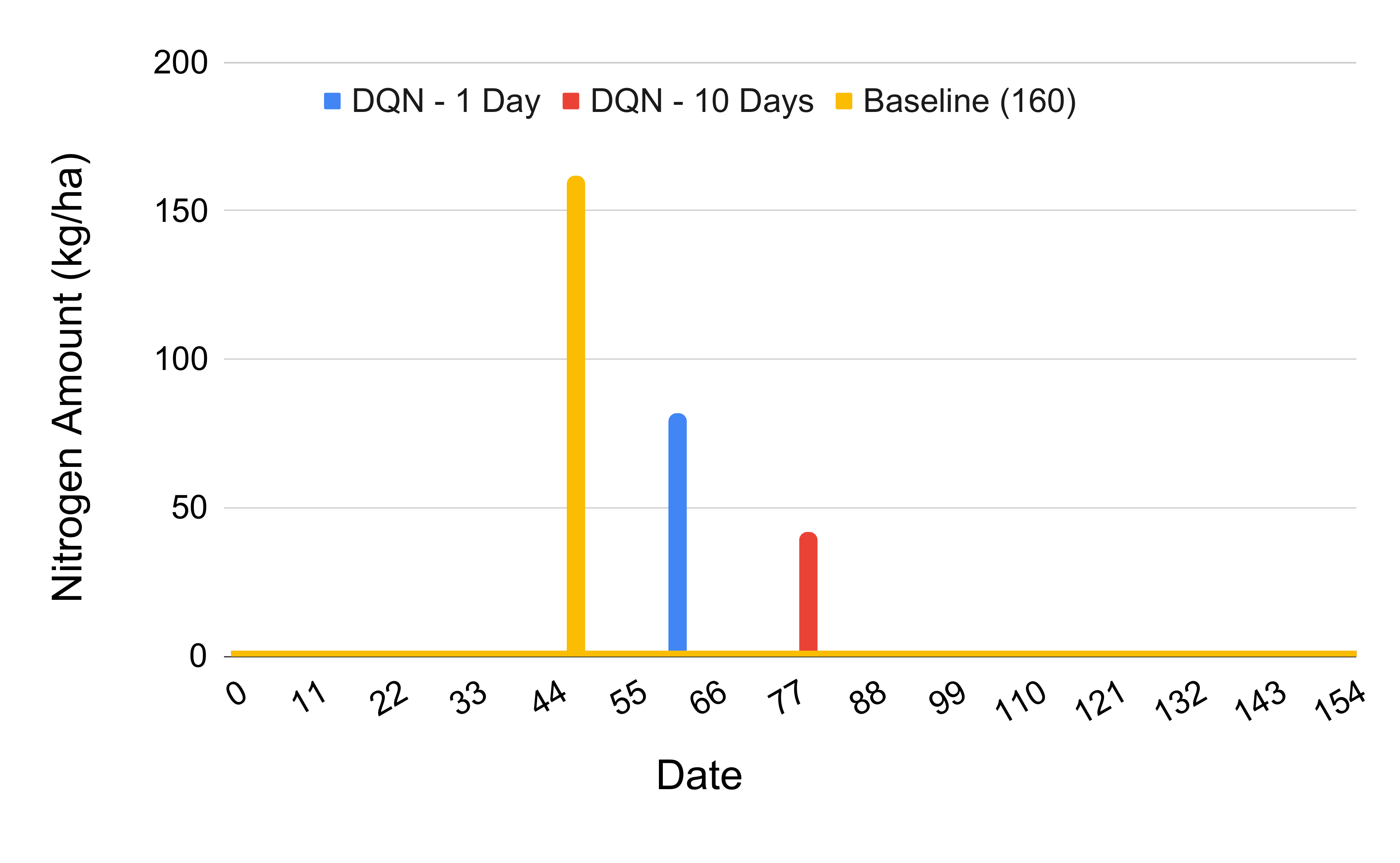}
   \caption{Applied actions under different action frequencies for Florida.}
   \label{fig: action florida}
\end{figure}

For Iowa, \cref{fig: action Iowa} shows the applied actions and Table~\ref{table:action frequency in Iowa} lists achieved cumulative reward and top weight at maturity under different actions. The DQN agents fertilized 5 to 6 times. Both the baseline method and the DQN policy with a 10-day action frequency  used 280 kg/ha nitrogen input. However, their results are relatively poor compared with the DQN policy with a 1-day action frequency.

The results for Florida are shown in \cref{fig: action florida} and Table~\ref{table:action frequency in Florida}. One can see that both baseline methods and our RL agent chose to fertilize once across the crop's life cycle. However, the RL agent tended to apply less nitrogen, as shown in \cref{fig: action florida}. According to Table~\ref{table:action frequency in Florida},  the performance of the DQN agent degrades a little bit under the reduced action frequency of 10 days, but is still better than that of the best baseline. 
\begin{table}[!ht]
    \centering
    \footnotesize
    \caption{Performance of trained policies under different action frequencies for Iowa. DQN (1 Day): RL agents are allowed to fertilize every day. DQN (10 Days): RL agents are only permitted to fertilize every ten days.
    }
    \begin{tabular}{l|l l}
    \hline
        Method & Top weight at maturity (kg/ha)  & Cumulative reward  \\ \hline \hline
        DQN (1 Day) & 21710.7 & 2147.1 \\ \hline
        DQN (10 Days) & 21700.6 & 2142.0 \\ \hline
        Baseline & 21709.5 & 2142.9 \\ \hline
    \end{tabular}

    \label{table:action frequency in Iowa}
\end{table}

\begin{table}[!ht]
    \centering
    \footnotesize
    \caption{Performance of trained policies under different action frequencies for Florida.}
    \begin{tabular}{l| l l}
    \hline
        Method & Top weight at maturity (kg/ha)  & Cumulative reward  \\ \hline     \hline
        DQN (1 Day) & 6310.8 & 619.8 \\ \hline
        DQN (10 Days) & 5728.9 & 565.7 \\ \hline
        Baseline & 5190.4 & 493.3 \\ \hline
    \end{tabular}

    \label{table:action frequency in Florida}
\end{table}

\begin{table*}[h]
    \centering
    \footnotesize
    \caption{State space description}
    \begin{tabular}{l|l|l}
    \hline
    State      & Description                                                                       & Included in Partial Observation Study?\\ \hline
    cumsumfert & cumulative nitrogen fertilizer applications (kg/ha)                               & \checkmark      \\ \hline
    dap        & days after simulation started                                                     & \checkmark      \\ \hline
    dtt        & growing degree days for current day (C/d)                                         & \checkmark      \\ \hline
    istage     & DSSAT maize growing stage                                                         & \checkmark      \\ \hline
    vstage     & vegetative growth stage (number of leaves)                                        & \checkmark      \\ \hline
    pltpop     & plant population density (plant/m2)                                               & \checkmark      \\ \hline
    rain       & rainfalls for the current day (mm/d)                                              & \checkmark      \\ \hline
    srad       & solar radiations during the current day (MJ/m2/d)                                 & \checkmark      \\ \hline
    tmax       & maximum temparature for current day (C)                                           & \checkmark      \\ \hline
    tmin       & minimum temparature for current day (C)                                           & \checkmark      \\ \hline
    nstres     & index of plant nitrogen stress (unitless)                                         &       \\ \hline
    pcngrn     & massic fraction of nitrogen in grains (unitless)                                  &       \\ \hline
    swfac      & index of plant water stress (unitless)                                            &       \\ \hline
    tleachd    & daily nitrate leaching (kg/ha)                                                    &       \\ \hline
    grnwt      & grain weight dry matter (kg/ha)                                                   &       \\ \hline
    cleach     & cumulative nitrate leaching (kg/ha)                                               &       \\ \hline
    cnox       & cumulative nitrogen denitrification (kg/ha)                                       &       \\ \hline
    tnoxd      & daily nitrogen denitrification (kg/ha)                                            &       \\ \hline
    trnu       & daily nitrogen plant population uptake (kg/ha)                                    &       \\ \hline
    wtnup      & cumulative plant population nitrogen uptake (kg/ha)                               &       \\ \hline
    xlai       & plant population leaf area index (m2\_leaf/m2\_soil)                              &       \\ \hline
    topwt      & top weight (kg/ha)                                       &       \\ \hline
    es         & actual soil evaporation rate (mm/d)                                               &       \\ \hline
    runoff     & calculated runoff (mm/d)                                                          &       \\ \hline
    wtdep      & depth to water table (cm)                                                         &       \\ \hline
    rtdep      & root depth (cm)                                                                   &       \\ \hline
    totaml     & cumulative ammonia volatilization (kgN/ha)                                        &       \\ \hline
    sw         & volumetric soil water content in soil layers (cm3 {[}water{]} /   cm3 {[}soil{]}) &       \\ \hline
\end{tabular}
\label{table:state-description}
\end{table*}

\section{Conclusion}


Effective nitrogen (N) management is crucial for maximizing crop yields and minimizing negative environmental impacts.
We present a framework for optimizing N management with deep reinforcement learning (RL) and crop simulations based on DSSAT. With the proposed framework, we train management policies with deep Q-network (DQN) and soft actor-critic (SAC) for the maize crop in Iowa and Florida, which are shown to outperform standard management practices. We also evaluate the effect of partial observation and reduced action frequencies.  
We believe our work demonstrates the potential of deep RL in optimizing crop management for more sustainable and resilient agriculture. 

Current study is focused on testing the proposed framework on a single type of crop in the simulator under a fixed weather condition (once the location is fixed). In the future, we will randomize weather during training and test the robustness of trained policies in the presence of uncertain weather conditions. 
{Moreover, we plan to leverage real-world data to calibrate the crop simulation models  and/or fine-tune the policy, which will help bridge the sim-to-real gap. Additionally, we plan to explore more advanced methods to improve the learning performance under partial observability. 
Besides, we plan to include other management practices such as irrigation and tillage in the action space. 
}
Finally, modifications can be made to Gym-DSSAT so that more types of crops can be tested with our proposed framework.

\section{Acknowledgment}
This work was supported by the C3.ai Digital Transformation Institute and NSF under the RI grant \#2133656. The authors thank Dr.~Bruno Castro da Silva for helpful discussions. The authors also thank Romain Gautron and Emilio Padrón González for the development of the Gym-DSSAT interface and helpful suggestions on its use in this work.
\newpage
{\small
\bibliographystyle{ieee_fullname}
\bibliography{refs.bib}
}






\end{document}